\documentclass[10pt,twocolumn,letterpaper]{article}

\usepackage[pagenumbers]{wacv} 
\usepackage{times}
\usepackage{epsfig}

\usepackage{graphicx}
\usepackage{amsmath}
\usepackage{amssymb}
\usepackage{booktabs}
\usepackage{algorithm}
\usepackage{algorithmic}
\usepackage{fontawesome}
\usepackage[dvipsnames]{xcolor}
\usepackage{amssymb}
\usepackage{pifont}
\usepackage{booktabs} 
\usepackage{color,soul} 
\usepackage{colortbl}
\usepackage{stfloats}
\definecolor{mygray}{gray}{.9}

\newcommand{\cmark}{\ding{51}}%
\newcommand{\xmark}{\ding{55}}%
\newcommand{\MoP}{MoP-CLIP }

\usepackage{fontawesome}
\def\faChecked{\FA\symbol{"F00C}}
\def\faCrossed{\FA\symbol{"F00D}}
\def\real{{\mathbb{R}}}

\def\zz{{\mathbf{z}}}
\def\xx{{\mathbf{x}}}
\def\yy{{\mathbf{y}}}
\def\cc{{\mathbf{c}}}
\def\pp{{\mathbf{p}}}

\newcommand{\mr}[1]{\mathrm{#1}}

\usepackage[pagebackref,breaklinks,colorlinks]{hyperref}

\usepackage[capitalize]{cleveref}
\crefname{section}{Sec.}{Secs.}
\Crefname{section}{Section}{Sections}
\Crefname{table}{Table}{Tables}
\crefname{table}{Tab.}{Tabs.}

\title{MoP-CLIP: A Mixture of Prompt-Tuned CLIP Models \\ for Domain Incremental Learning}

\begin{document}


\author{Julien Nicolas\\
ETS Montreal\\
\and
Florent Chiaroni\\
Thales Digital Solutions\\
\and
Imtiaz Ziko\\
Thales Digital Solutions\\
\and
Ola Ahmad\\
Thales Digital Solutions\\
\and
Christian Desrosiers\\
ETS Montreal\\
\and
Jose Dolz\\
ETS Montreal\\
}

\maketitle

\begin{abstract}
    Despite the recent progress in incremental learning, addressing catastrophic forgetting under distributional drift is still an open and important problem. Indeed, while state-of-the-art domain incremental learning (DIL) methods perform satisfactorily within known domains, their performance largely degrades in the presence of novel domains. This limitation hampers their generalizability, and restricts their scalability to more realistic settings where train and test data are drawn from different distributions. 
    To address these limitations, we present a novel DIL approach based on a mixture of prompt-tuned CLIP models (MoP-CLIP), which generalizes the paradigm of S-Prompting to handle both in-distribution and out-of-distribution data at inference. 
    In particular, at the training stage we model the features distribution of every class in each domain, learning individual text and visual prompts to adapt to a given domain. At inference, the learned distributions allow us to identify whether a given test sample belongs to a known domain, selecting the correct prompt for the classification task, or from an unseen domain, leveraging a mixture of the prompt-tuned CLIP models. 
    Our empirical evaluation reveals the poor performance of existing DIL methods under domain shift, and suggests that the proposed \MoP performs competitively in the standard DIL settings while outperforming state-of-the-art methods in OOD scenarios. These results demonstrate the superiority of \MoP, offering a robust and general solution to the problem of domain incremental learning.

\end{abstract}

\section{Introduction}
In machine learning, it is a common practice to assume that both training and test data follow the same underlying distribution. In real-world scenarios, however, this strong assumption is rarely met, leading to substantial performance degradation when the trained model is evaluated on test samples under a distributional drift. A simple solution to alleviate this issue is to train the model on the labeled samples from the new domain. However, when the learning is performed in a sequential manner on multiple domains, contemporary deep learning models tend to suffer from the phenomenon of \textit{catastrophic forgetting}, wherein the acquired knowledge from previous domains is typically erased.

A simple strategy to address this issue consists in training different models, one per single domain. However, this approach is suboptimal, as all these models must be stored for future usage and the domain identity is not necessarily known at test time. To tackle the issue of forgetting learned knowledge, domain incremental learning (DIL) has recently emerged as an appealing alternative that alleviates the need to store multiple domain-specific networks. Among the different DIL approaches, rehearsal \cite{aljundi2019gradient,bang2021rainbow,hou2019learning,rolnick2019experience} and distillation-based \cite{ahn2021ss,hou2018lifelong,lee2019overcoming} methods, which leverage a buffer of stored exemplars from old domains, dominate the literature. Nevertheless, from a privacy and storage standpoint, \textit{exemplar-free} DIL approaches may offer a better solution in practical settings. 

\begin{figure}[h!]
    \centering
    \includegraphics[width=1.0\linewidth]{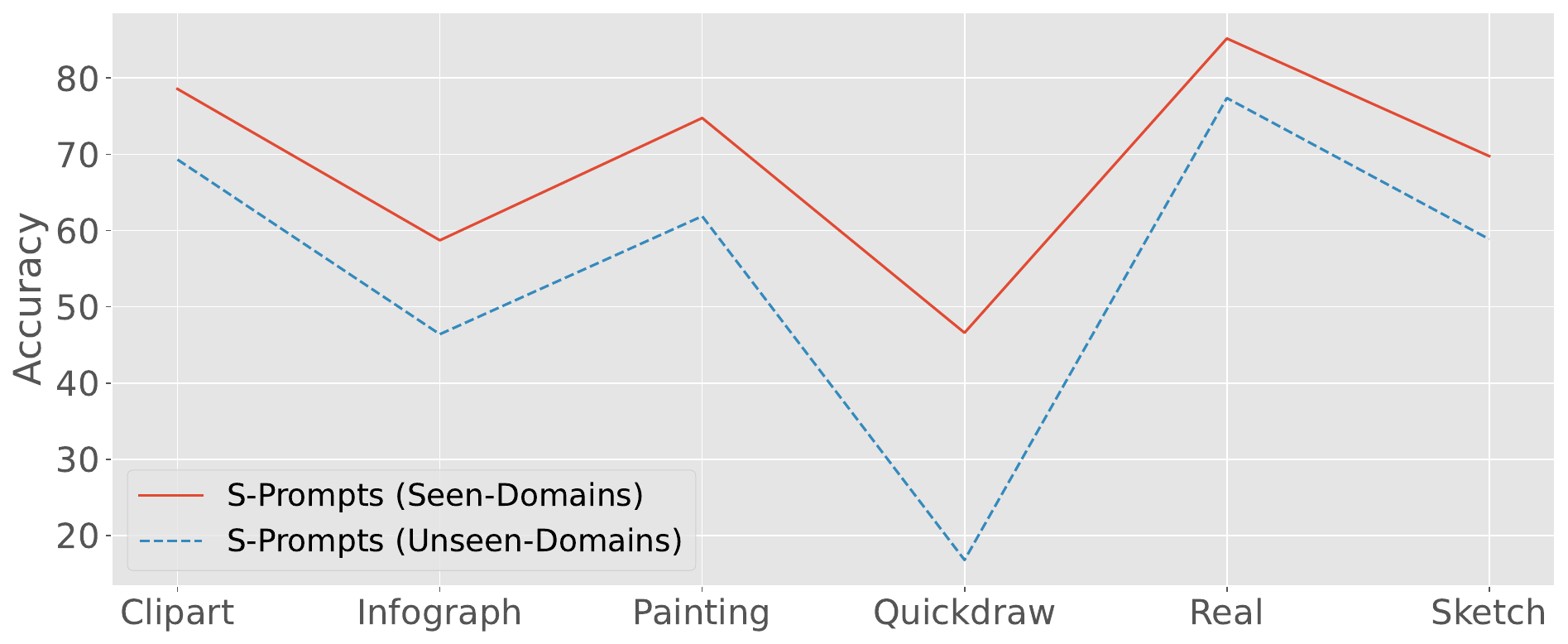}
    \caption{\textbf{Performance degradation under the presence of domain shift} between adaptation and testing samples, which shows that sota DIL approaches do not generalize well. We employ S-Prompts \cite{wang2022sprompts} as use-case. The \textcolor{red}{red line} represents the performance across each test domain, when all domains have been seen by the model. In contrast, the \textcolor{blue}{blue dotted} line shows the performance of the same model when the test domain remains unknown, highlighting the performance degradation under distributional shift.}
    \label{fig:motivation}
\end{figure}

An appealing alternative to mitigate knowledge forgetting is prompt-learning, which is driving progress in a wide span of transfer learning problems \cite{le2021many,zhou2022conditional}. 
In this approach, domain-specific knowledge is preserved in the form of textual and visual prompts, alleviating the need of storing exemplars per domain. While some methods advocate for the joint learning of prompts across tasks \cite{douillard2022dytox,wang2022learning}, the recent work in \cite{wang2022sprompts} instead favors the learning of the prompts independently, suggesting that this leads to the best performance per domain. This learning paradigm, referred to as S-Prompting \cite{wang2022sprompts}, circumvents the issue of using expensive buffers by optimizing per-domain prompts, which are leveraged at testing time. In particular, centroids for each domain are obtained during training by applying K-Means on the training image features, which are generated with the fixed pre-trained transformer without using any prompts. Then, during inference, the standard KNN algorithm is used to identify the nearest centroid to the test image, whose associated domain prompt is added to the image tokens for classification. 
Despite the empirical performance gains observed by these approaches \cite{douillard2022dytox,wang2022learning,wang2022sprompts}, 
a current limitation hampering their generalization is that they perform satisfactorily in \textit{known} domains, but typically fail when \textit{unseen} domains are presented (see Fig.~\ref{fig:motivation}). This is particularly important in real-world scenarios where training and testing data of the \textit{a priori} same domain may present distributional drifts that degrade the model performance. In the case of S-Prompts \cite{wang2022sprompts}, we argue that a potential reason behind this suboptimal performance stems from forcing the model to select a single domain (i.e., the closest one), which might be indeed far in the feature space.

Motivated by these limitations, we introduce a novel \textit{exemplar-free} DIL solution, based on prompt learning, which generalizes the recent S-liPrompts approach \cite{wang2022sprompts} for both in-distribution and out-of-distribution data. Specifically, our contributions can be summarized as follows:

\begin{itemize}
    \item We first expose that existing state-of-the-art domain incremental learning approaches suffer in the presence of distributional shift between samples used for adaptation and testing, which hampers their generalization to unseen domains (Fig.~\ref{fig:motivation}).
    \item Based on these observations, we present a novel DIL strategy based on a mixture of prompt-tuned (MoP) CLIP models, generalizing the recent S-liPrompts approach \cite{wang2022sprompts} to work with both in-distribution and out-of-distribution data. In particular, the proposed approach learns class-wise features distributions for each domain, allowing to detect whether a given sample comes from a known domain.     
    \item The proposed approach is \textit{exemplar-free}, reducing the computational burden compared to conventional methods, and \textit{agnostic to the sequence order}. 
    \item Extensive experiments demonstrate that our approach performs at par with state-of-the-art DIL methods on known domains, while largely outperforming them under distributional drifts.
\end{itemize}

\section{Related Work}

\noindent \textbf{Domain-Incremental learning (DIL)} 
refers to continual learning scenarios in which the distribution of instances from fixed classes changes between domains. 
These real-world scenarios include, for example, the recognition of objects where new instances from varying environments appear in each new domain \cite{core50}, or autonomous driving, where the car is exposed to ever-changing conditions. 
We focus on the domain-agnostic scenario, where the sample's domain remains unknown at inference time. The major challenge of this task is to find a good trade-off to adapt to the new instances distribution without deteriorating performance for samples of the previous distributions (i.e., alleviating \textit{catastrophic forgetting}).
The literature on this subject is abundant, where the main approaches are based on weight regularization \cite{kirkpatrick2017overcoming, zenke2017continual, chaudhry2018riemannian}, knowledge distillation in a teacher-student setting using current examples \cite{li2017learning} or a memory buffer \cite{chaudhry2019tiny} and methods using or generating latent features \cite{shin2017continual, pellegrini2020latent} or gradient examplars \cite{lopez2017gradient,chaudhry2019tiny, marra2019incremental}. Nevertheless, these approaches require the use of \textit{exemplars} from seen domains, which may result in storage, security and privacy issues. In contrast, the proposed approach only requires the storage of a single prototype per class and domain, which largely alleviates these issues. 

\vspace{2pt}\noindent \textbf{Prompt learning.} Driven by the advances in Natural Learning Processing, prompt learning has emerged as an appealing learning strategy to adapt large scale pre-trained models to downstream tasks. While initial attempts to adapt language-vision models have centered on carefully designing handcrafted prompts \cite{brown2020language}, recent works focus on optimizing a task-specific continuous vector, which is optimized via gradients during fine-tuning \cite{zhou2022learning,zhou2022conditional,lu2022prompt,ju2022prompting}. An underlying limitation of these approaches arises from the inherent disparity between language and vision modalities, and thus fine-tuning only text prompts for visual recognition tasks may yield suboptimal performance. Motivated by this, visual prompt tuning (VPT) \cite{jia2022visual} was proposed as a powerful alternative to text prompting. In this approach, authors propose to optimize task-specific learnable prompts in either the input or visual embedding space. Following the satisfactory results achieved by VPT, fine-tuning visual prompts has gained popularity recently, particularly for adapting pre-trained models to novel unseen categories \cite{sohn2023visual,chen2023understanding,xing2022class,xing2022class}.


\vspace{2pt}\noindent \textbf{Prompt tuning in domain incremental learning.} This paradigm protects against catastrophic forgetting by optimizing a small set of learnable prompts. This contrasts with classical approaches which modify all the network parameters (or a subset),or store \textit{exemplars} in a buffer. Despite the success observed in other tasks, the literature on prompt tuning for domain incremental learning remains underexplored, with just a handful works addressing this problem \cite{wang2022sprompts,wang2022learning,douillard2022dytox}. For example, S-Prompts \cite{wang2022sprompts} learns in isolation a set of prompts per domain, and dynamically selects which set to use at test-time using a fixed key/value dictionary where the keys are computed with K-Means and the values represent the sets of prompts. L2P \cite{wang2022learning} uses an incrementally learnable key/value mechanism to select which prompts to prepend to the input image tokens at test-time, hence breaking the isolation between domains, which contrasts with our work, as it learns domain prompts independently. 
A main difference with these, and conventional DIL approaches, is that the proposed approach explicitly tackles the generability performance in domain incremental learning, while maintaining at par accuracy in known domains, which remains underexplored.

\begin{figure*}[t]
    \centering
    \includegraphics[width=0.85\linewidth]{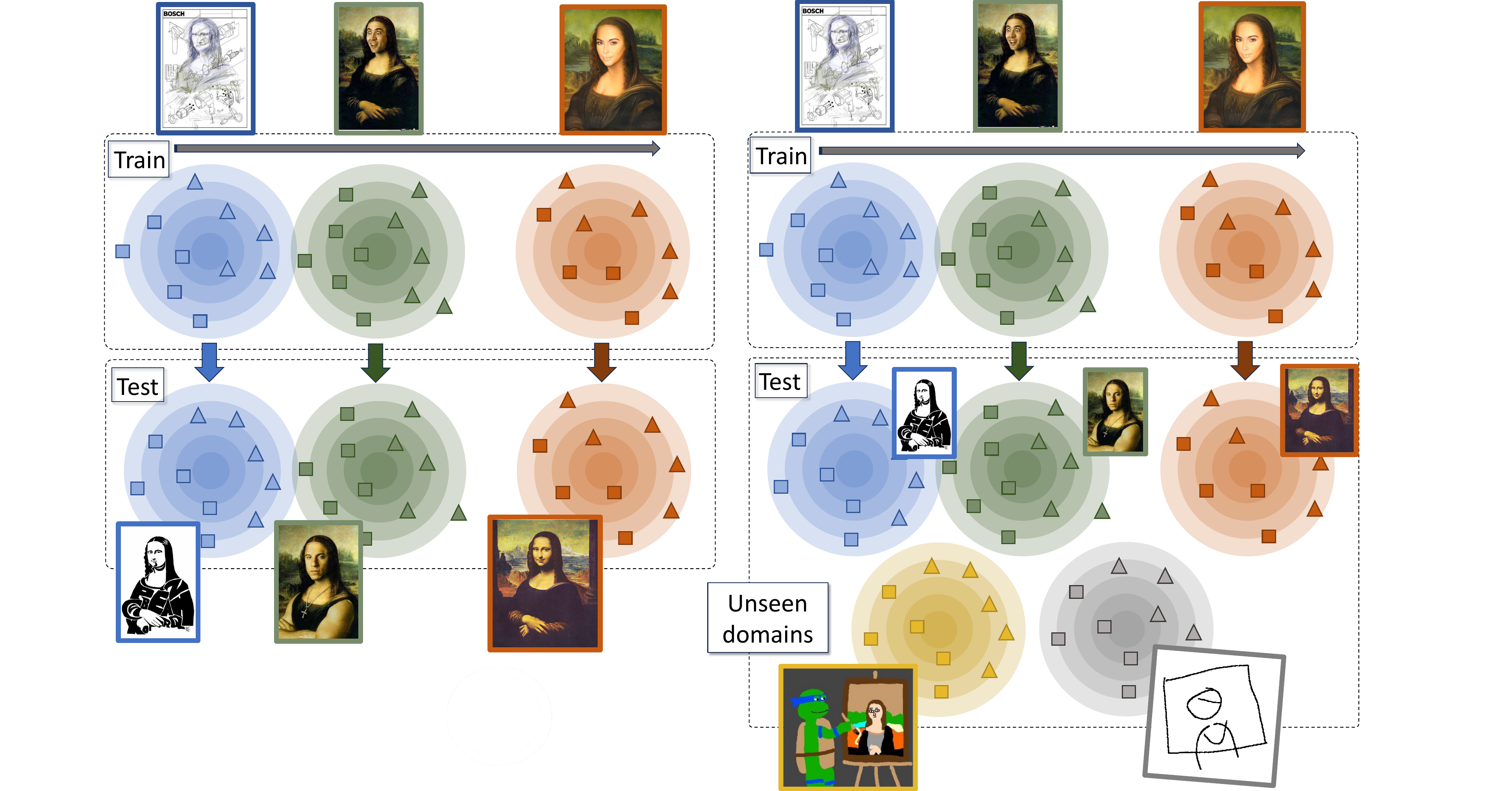}
    \caption{\textbf{Proposed generalization scenario for domain incremental learning} Standard problem (\textit{left}): Only in-domain examples are encountered at test time. Addressed problem (\textit{right}): Both in-domain and out-of-domain examples are presented at test time.}
    \label{fig:testing-scenario}
\end{figure*}

\vspace{2pt}\noindent \textbf{Domain generalization (DG)} Existing literature on DG strongly relies on supervised knowledge from source domain data, regardless of whether it originates from a single domain \cite{wang2021learning} or multiple domains \cite{yao2022pcl,zhang2022exact,zhang2023nico++,chen2022compound}, which may not be realistic in continually changing scenarios, as knowledge comes in a sequential manner. Additionally, in scenarios involving distributional shifts, DG approaches primarily focus on the target domain, increasing the potential risk of catastrophic forgetting on previously learned domains \cite{liu2022deja}.

\begin{figure*}[t]
\centering
\includegraphics[width=0.98\textwidth]{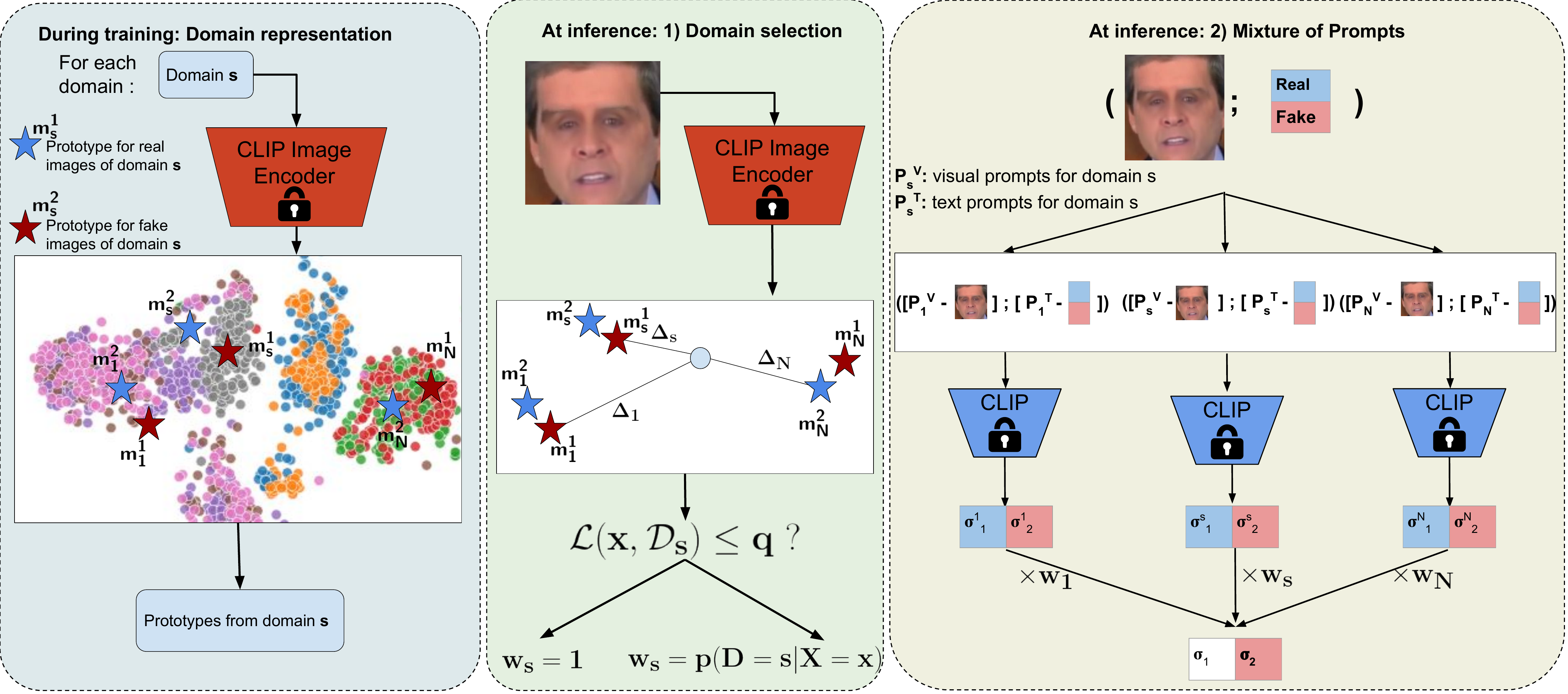}
\caption{\textbf{Overview of MoP-CLIP.} The training phase (\textit{left}): class-wise prototypes are identified from in-distribution domains. Inference (\textit{middle} and \textit{right}): domain selection and ensembling (Mixture of Prompts), respectively, for in-distribution and out-of-distribution samples. For simplicity, we depict the pipeline for 2 classes (Real \textit{vs} Fake). However, the procedure for multiple classes (e.g., DomainNet or CoRE50) is exactly the same.}
\label{inference}
\end{figure*}

\section{Method} 
An overview of MoP-CLIP is illustrated in Fig.~\ref{inference}, which contains two phases: \textit{i)} learning of in-distribution domain-specific visual and text prompts (sec.~\ref{sec:PL}) and \textit{ii)} selection of optimal prompts for a given test sample (sec.~\ref{sec:infer}). 

\subsection{Problem definition}
Let us denote as $\mathcal{S} = \left \{\mathcal{D}_{s} \right \}_{s=1}^{N}$ the sequence of datasets presented to the model in our incremental learning scenario, with $N$ being the final number of domains. Each dataset is defined as $\mathcal{D}_s \!=\! \left \{ \xx_i^s,\yy_i^s \right \}_{i=1}^{|\mathcal{D}_s|}$, where  $\xx_i\in\real^{W\times H \times C}$ represents an image of size $W\!\times\!H$ and $C$ channels, and $\yy_i \!\in\!\{0,1\}^{K}$ is its corresponding one-hot label for $K$ target classes. In this setting, we have access to only one domain $\mathcal{D}_s$ at a time and storing samples from previous seen domains, commonly referred to as \textit{exemplars}, is not allowed. 
Each time a new domain $\mathcal{D}_s$ becomes accessible, DIL aims to improve the model's performance on $\mathcal{D}_s$, while avoiding the loss of knowledge for past domains, $\mathcal{D}_{s-1}, \mathcal{D}_{s-2}, ... \mathcal{D}_{1}$. In the proposed setting, and in contrast to most existing literature on DIL, we assume that the model should also generalize well on unseen datasets, i.e.,  $\mathcal{D}_{s+1}, \mathcal{D}_{s+2}, ..., \mathcal{D}_{|\mathcal{D}_s|}$ (Fig. \ref{fig:testing-scenario}). In other words, our learning scenario leverages \textit{backward transfer} to avoid catastrophic forgetting on seen domains, while optimizing \textit{forward transfer} to facilitate knowledge transfer to new tasks/domains. Our motivation behind this bi-directional performance assessment relies on the realistic assumption that a distributional drift between training and testing data always exists. 

\subsection{Prompts Learning}
\label{sec:PL}

Following the setting in \cite{wang2022sprompts}, we define $f_{\theta}$ as the pre-trained vision transformer that generates a visual embedding $\zz^v\!=\!f_{\theta}(\xx_{\mr{tok}}) \in \real^L$, where $\xx_{\mr{tok}}\in\real^{WH/R^2\times M^v }$ corresponds to the image tokens (or patches), $WH/R^2$ is the number of tokens, $R$ is the width/height of the (square) patch and $M^v$ is the dimension of the image tokens embedding. We also define $f_{\phi}$, a pre-trained text transformer that generates text embeddings of dimension $M^t$ from class names tokens $\cc_k$ for $k\!\in\!\{1,...,K\}$.
For each new domain $\mathcal{D}_s$ in the sequence $\mathcal{S}$, we can adapt the model by learning a visual prompt $\pp_s^v\!\in\!\real^{L^v \times M^v}$ and a text prompt $\pp_s^t \in \real^{L^t \times M^t}$, following \cite{wang2022sprompts}. In particular, these prompts are a set of continuous learnable parameters, where $L^v,L^t$ are the visual and text prompt length.
Thus, for the set of domains $\mathcal{S}$, we have a set of domain-specific visual and text prompts, denoted as $\mathcal{P}^v\!=\!\{\pp^v_1,...,\pp^v_{N}\}$ and  $\mathcal{P}^t\!=\!\{\pp^t_1,...,\pp^t_{N}\}$. Now, with the domain-specific prompts, we can modify the embeddings that will be provided to the visual and text encoders, $f_{\theta}$ and $f_{\phi}$. Concretely, for an image of domain $s$ and class $k$, the input of the visual transformer is defined as $\tilde{\xx}^v=[\xx_{\mr{tok}}, \pp^v_s, \xx_{\mr{cls}}]$ with $\xx_{\mr{cls}}$ the classification token of the ViT. Similarly, the input of the text transformer is defined as $\tilde{\cc}^{t}_{k}=[\pp^t_s, \cc_k]$. We then denote as $\tilde{\zz}^v\!=\! f_{\theta}(\tilde{\xx}^v)$ and $\tilde{\zz}^t_k\!=\! f_{\phi}(\tilde{\cc}^{t}_k)$ the embeddings of these inputs. 
The posterior probability of a given image $\xx_i$ from $\mathcal{D}_s$ belonging to class $k$ can be therefore defined as:
\vspace{-1mm}
\begin{equation}
p(\yy_k|\xx, s)=\frac{e^{\cos(\tilde{\zz}^v, \tilde{\zz}^t_k)}}{\sum_{j=1}^K e^{\cos(\tilde{\zz}^v, \tilde{\zz}^t_j)}},
\label{eq:posterior}
\vspace{-1mm}
\end{equation}
where $\cos(\mathbf{a},\mathbf{b})=\frac{\mathbf{a}\cdot \mathbf{b}}{\|\mathbf{a}\| \, \|\mathbf{b}\|}$ is the cosine similarity between vectors $\mathbf{a}$ and $\mathbf{b}$.

\subsection{Inference}
\label{sec:infer}
At test time, the domain of the images to classify remains unknown. In S-liPrompts \cite{wang2022sprompts}, the domain $s^*$ closest to a given test sample is selected by finding the minimum distance between the visual embeddings and prototypes computed with K-Means over the domains $\mathcal{S}$. 
This strategy is generally effective in finding the closest domain when $\xx \in \mathcal{D}_s$ and $\mathcal{D}_s$ has been already presented to the model. In this setting, $p(\yy_k|\xx, s)$ yields satisfying predictions, as the domain of the sample $\xx$ can be easily inferred and the scenario becomes a classification task under in-distribution data. Nevertheless, when the model has not been exposed to $\mathcal{D}_s$ during training or adaptation, the selection of an existing closest domain (other than $\mathcal{D}_s$) might not match with the real distribution of the new domain. In this case, the strategy used in S-liPrompts may actually move the test sample away from its original distribution. 
To overcome this issue, we propose to enhance the domain selection mechanism in two separate ways: \textit{i)} dynamically allowing the model to select $n$ close domains and \textit{ii)} leveraging per-domain predictions in an ensembling scheme for samples of unseen domains.

To select the right prompt, we propose a strategy based on a set of class-specific prototypes for each domain, $\mathcal{E}_s\!=\!\{\boldsymbol{m}_s^k\}_{k=1}^K$, instead of prototypes obtained with K-Means as in \cite{wang2022sprompts}. Let $\mathcal{D}_s^k \subset \mathcal{D}_s$ be the samples of domain $\mathcal{D}_{s}$ belonging to the class $k$, we compute the the prototype of class $k$ for domain $\mathcal{D}_s$ by averaging the visual embeddings of examples in $\mathcal{D}_s^k$:
\begin{equation}
    \boldsymbol{m}_s^k = \frac{1}{|\mathcal{D}_s^k|} \sum_{\{ \boldsymbol{z}^v\,|\,\xx \in \mathcal{D}_s^k\}} \!\!\!\!\!\boldsymbol{z}^v
\end{equation}
Next, we present how these prototypes are used to select the domain and how they are leveraged in our approach. 
\label{domain_select}
\paragraph{\textbf{i) Domain Selection.}} Given the class-specific prototypes, we select the domain $s^*$ of a test example $\xx$ as the one with the nearest prototype for any class:
\begin{equation}
\label{eq:domainselection}
s^* \, = \, \operatorname*{argmin}_{1\leq s \leq N} \Delta_s(\xx)
\end{equation}
with 
\begin{equation}
\Delta_s(\xx) \, = \, \operatorname*{min}_{\boldsymbol{m}_s^k \in \mathcal{E}_s} \|\zz^v - \boldsymbol{m}_s^k\|_{2}.
\end{equation}

As mentioned before, test examples may also come from an out-of-distribution (OOD) domain (i.e., not part of any domains encountered at training time). To determine if a given sample $\xx$ is from a previously-seen domain or is OOD, we compare its distance to the closest prototype of the selected domain, $\Delta_{s^*}(\xx)$, with the distances of training samples from that domain. Let $\Psi^k_{s} \, = \, \big\{\| \zz^v - \boldsymbol{m}_{s}^k\|_2 \, | \, \xx \in \mathcal{D}_{s}^k\big\}$ be the set of distances for domain $\mathcal{D}_{s}$ and  class $k$. During training, the distribution of distances for each domain $\mathcal{D}_{s}$ and class $k$ is estimated from $\Psi^k_{s}$ with a Gaussian of mean $\mu^k_s$ and standard deviation $\sigma^k_s$.

At test time, we find the class corresponding the nearest prototype for the selected domain, i.e., $k^* = \operatorname*{argmin}_{1\leq k \leq K} \|\zz^v - \boldsymbol{m}_{s^*}^k\|_{2}$. We then use the distribution $P\!=\! \mathcal{N}(\cdot\,; \mu^{k^*}_{s^*}, \sigma^{k^*}_{s^*}$) to determine whether $\Delta_{s^*}(\xx)$ is normal. Specifically, we classify a sample $\xx$ as in-distribution if $F(\Delta_{s^*}(\xx)) \leq q$ where $F$ is the cumulative distribution function of $P$, i.e., $F(x)\!=\!P(X \leq x)$ and $q$ is a specified threshold.  

\noindent Afterwards, if $\xx$ is in-distribution, we use $p(\yy_k\,|\,\xx,  s^*)$ to classify $\xx$. Otherwise, $\xx$ belongs to a new (unseen) domain. In such case, we propose the following ensembling technique to classify it.
\paragraph{\textbf{ii) Ensembling}}

If $\xx \in \mathcal{D}_{s'}$ and $\mathcal{D}_{s'}$ has not been encountered during training, we model $\zz^v$ as being part of a mixture of the known domains. In particular, we resort to a Gaussian mixture model to estimate the mixture weights ($w_s = p(s|\xx)$). While this could be done with $L$-dimensional covariance and mean vectors per domain (on the features), it does not perform well as $L$ increases. We propose the following model: 

\begin{equation}
\begin{split}
 w_s & = p(\xx \in \mathcal{D}_{s})\\
    & = \frac{\mathcal{N}(\Delta_s(\xx);\mu^{k^*}_{s}, \sigma^{k^*}_{s})}{\sum_j \mathcal{N}(\Delta_j(\xx);\mu^{t^*}_{j}, \sigma^{t^*}_{j})}, \\
\end{split}
\label{ensembling}
\end{equation}
where $t^* = \operatorname*{argmin}_{1\leq k \leq K} \|\zz^v - \boldsymbol{m}_{j}^k\|_{2}$. Note that the hypotheses done to reach the proposed model in eq. (\ref{ensembling}) are detailed in Supplemental Material. 
We then combine the predictions using the different prompts ($p(\yy_k|\xx,  s)$) based on those weights: 
\begin{equation}
p(\yy_k|\xx) = \sum_{s=1}^{N} p(\yy_k|\xx,  s) \cdot w_s 
\end{equation}


\section{Experiments}

The experiments reported in this section validate empirically that MoP-CLIP yields competitive performance compared to state-of-the-art DIL when dealing with in-domain (ID) examples, while significantly outperforming these approaches in the presence of out-of-domain (OOD) examples. Furthermore, we perform a series of ablation experiments to better identify the impact of the key components of the proposed method. 

\subsection{Experimental setup}

\noindent {\textbf{A. Datasets.}}
To assess the performance of the proposed method, we resort to three popular DIL benchmarks which have been extensively used in the literature: 
CDDB-Hard \cite{li2023continual}, DomainNet \cite{domainnet}, and CORe50 \cite{core50}, whose details are given below:

\noindent \textbf{CDDB Dataset} \cite{li2023continual} is a continual (incremental) deepfake detection benchmark, whose goal is to identify real and fake images across different domains. In particular, in the proposed work we employ the Hard setting as in \cite{wang2022sprompts}, which is the most challenging track of CDDB. This dataset contains a total of 27,000 images across 5 different domains: GauGAN, BigGAN, WildDeepfake, WhichFaceReal, and SAN. We also use Glow, StarGAN and CycleGAN to evaluate OOD performance.
    
\noindent  \textbf{DomainNet} \cite{domainnet} is a dataset for domain adaptation commonly used to benchmark DIL methods. It contains a total of 600,000 images across 6 different domains, each containing the same 345 classes. In particular, we use the  experimental setup presented in CaSSLe \cite{fini2022self}.

\noindent \textbf{CORe50} \cite{core50} is a dataset designed for continual object recognition. However, in this work we focus on its domain-incremental learning scenario. This setting is comprised of 11 distinct domains, each containing the same 50 object categories. From the 11 domains, 8 are composed of 120,000 images which are seen sequentially during training, whereas the remaining 3 domains compose the fixed unseen test set.
\paragraph{\textbf{B. Comparison methods.}} We benchmark MoP-CLIP to several state-of-the-art DIL methods. These include \textbf{non-prompting} approaches (EWC \cite{kirkpatrick2017overcoming}, LwF \cite{li2017learning}, ER \cite{chaudhry2019tiny}, GDumb~\cite{prabhu2020gdumb}, BiC~\cite{wu2019large}, DER++~\cite{buzzega2020dark} and Co$^2$L~\cite{cha2021co2l}), \textbf{prompting-based} methods (L2P \cite{wang2022learning}, DyTox \cite{douillard2022dytox} and S-lilPrompts \cite{wang2022sprompts}) and a \textbf{self-supervised} learning method, CaSSLe \cite{fini2022self}, following the experimental set-up in \cite{wang2022sprompts}. For OOD experiments, we only evaluate those methods that are in direct competition with our approach, in terms of \textit{exemplars} buffer use. In particular, we compare to the following methods, whose respective codes are publicly available: EWC\footnote{\url{https://github.com/G-U-N/PyCIL/}}, LwF\footnote{\url{https://github.com/G-U-N/PyCIL/}}, DyTox\footnote{\url{https://github.com/arthurdouillard/dytox}}, L2P\footnote{\url{https://github.com/JH-LEE-KR/l2p-pytorch}}, and S-liprompts\footnote{\url{https://github.com/iamwangyabin/S-Prompts}}.

\paragraph{\textbf{C. Evaluation metrics and protocol.}} 
To assess the performance of the proposed approach, we resort to standard metrics in the incremental learning literature.
\textbf{In-domain setting:} On DomainNet and CDDB-Hard we follow the original work in \cite{li2023continual} and employ the average classification accuracy (AA), as well as the average forgetting degree (AF), which is the mean of the popular backward transfer degradation (BWT). We formally define the average accuracy as $AA = \frac{1}{N}\sum_{i=1}^{N} A_{i,N}$ with $A_{i,N}$ the accuracy on domain $i$ measured after having trained on $N$ domains. This metric is computed at the end, i.e., after having seen all the domains, e.g., on CDDB: GauGAN $\rightarrow{}$BigGAN$\rightarrow{}$ WildDeepfake$\rightarrow{}$WhichFaceReal$\rightarrow{}$SAN.  
Furthermore, the average forgetting degree on CDDB can be defined as $\frac{1}{N-1}\sum_{i=1}^{N-1} BWT_{i}$ with $BWT_{i} = \frac{1}{N-i-1}\sum_{j=i+1}^{N} (A_{i,j} - A_{i,i})$ as originally proposed in \cite{li2023continual} (i.e., the forgetting degree is computed for each domain at each adaptation step, then averaged).
\textbf{Out-of-domain setting:} 
We follow \cite{core50} to compute the AA 
on CORe50 on the fixed test set, which contains 3 hold-out splits that can be considered as OOD with respect to the training set. Furthermore, as in \cite{wang2022sprompts}, we compute the AA 
on 3 unseen domains (Glow, StarGAN and CycleGAN) in CDDB-Hard.
Last, as no independent hold-out subset of unseen domains exists for DomainNet, we propose using the Cumulative Accuracy on the unseen domains during the incremental learning of the model (i.e., average accuracy on the unseen domains averaged on all the steps), defined as follows: $CA = \frac{1}{N-1}\sum_{i=1}^{N-1} \frac{1}{N-j-1}\sum_{i=j}^{N} A_{i,j}$. 
\paragraph{\textbf{D. Implementation details}}
We use the same setting as \cite{wang2022sprompts}, i.e. use ViT-B/16 \cite{dosovitskiy2020image} as our base image encoder and the text encoder of CLIP, both initialized by CLIP pretraining on ImageNet \cite{imagenet}. We follow \cite{wang2022sprompts} and use the same image encoder model as a backbone (i.e., ViT-B/16 \cite{dosovitskiy2020image} pretrained on ImageNet \cite{imagenet}) across all the compared methods, for a fair comparison. 
As suggested in \cite{wang2022sprompts}, we use a more advanced backbone (i.e. ConViT pretrained on ImageNet \cite{imagenet}) on DyTox \cite{douillard2022dytox} as it underperforms a random model with ViT-B/16 as backbone. 
We empirically fix $q = 0.94$ for the 3 datasets, based on the ablation study in Figure \ref{fig:ablation-study}, such that we do not deteriorate ID performance while improving OOD performance on CDDB-Hard. 
For EWC, LwF and CaSSLe, we use the same hyperparameters as in the original papers, whereas we keep the hyperparameters reported in \cite{wang2022sprompts} for DyTox, L2P and S-Prompts. 

\subsection{Results}
\noindent \textbf{In-domain distributions.} 
We first evaluate the proposed approach in the standard DIL scenario where the testing samples are drawn from the same distribution as the training/adaptation images. These results, which are reported under the \textit{Seen-Domains} columns of Tables \ref{table:cddbresults} and \ref{table:domainnetresults}, demonstrate that the proposed \MoP approach yields superior performance than existing \textit{exemplar-free} methods. In particular, \MoP outperforms the very recent approaches DyTox \cite{douillard2022dytox} and L2P \cite{wang2022learning} by large margin, with improvement gains of around 20-30\% in terms of average classification accuracy under the same storage conditions. Furthermore, the degree of knowledge forgetting is also largely reduced, going from -45.85 in DyTox to -0.79 in our approach. Furthermore, if storing exemplars is allowed, DyTox  \cite{douillard2022dytox} significantly improves its performance, but still underperforms our approach yet incurring a non-negligible overhead. 
Last, it is noteworthy to highlight that the proposed approach reaches similar performance than S-liPrompts \cite{wang2022sprompts} in this scenario, with at par values in the CDDB-Hard dataset and remarkable performance gains in DomainNet. Note that this result is somehow expected, as our approach is a generalization of S-liPrompts for the OOD scenario, and differences in the in-distribution setting may come from the domain prompt selected. 

An interesting observation is that prompting-based methods, which do not store exemplars from old tasks, typically outperform their buffer-storage counterparts. 
For example, S-liPrompts \cite{wang2022sprompts} and \MoP bring considerable improvements compared to LUCIR (between 6-8\%) or iCaRL (ranging from 9 to 15\%). We hypothesize that this phenomenon comes from the absence of interference between domains when doing the adaptation. In this scenario, the knowledge from previously learned domains remains isolated in the form of optimized domain prompts, and the only knowledge shared is derived from pre-trained transformers.

\begin{table}[h!]
\caption{\textbf{Results on CDDB-Hard for both ID and OOD scenarios.} Evaluation of existing state-of-the-art DIL methods in the standard \textit{seen-domain} setting and more challenging \textit{unseen-domain} scenario. For the unseen-domain experiments, we only reproduced the results for related (i.e., \textit{exemplar-free}) methods. Best results are highlighted in \textbf{bold}.}
\vspace{-0.6cm}
\label{table:cddbresults}
\begin{center}
\resizebox{0.99\linewidth}{!}{
\begin{tabular}{lccccc}
\toprule 
  & & & \multicolumn{2}{c}{Seen-Domains} & Unseen-Domains \\
 Method & Prompts & Buffer size & AA ($\uparrow$) & AF ($\downarrow$) & AA ($\uparrow$) \\
\midrule
 LRCIL$_\text{ IROS'20}$\cite{pellegrini2020latent} & \xmark & & 76.39 & -4.39 & -   \\ 
 iCaRL$_\text{ WIFS'19}$\cite{marra2019incremental} &  \xmark&\textit{100ex/class} & 79.76 & -8.73  & - \\
 LUCIR$_\text{ CVPR'19}$\cite{hou2019learning} & \xmark &  & 82.53 & -5.34 & -  \\
\midrule
   LRCIL$_\text{ IROS'20}$\cite{pellegrini2020latent}  & \xmark&  & 74.01 & -8.62& -  \\ 
 iCaRL$_\text{ WIFS'19}$\cite{marra2019incremental} & \xmark & \textit{50ex/class} & 73.98 & -14.50& -  \\
 LUCIR$_\text{ CVPR'19}$\cite{hou2019learning} &  \xmark&  & 80.77 & -7.85& -  \\
 DyTox$_\text{ CVPR'22}$\cite{douillard2022dytox} & \cmark & & 86.21 & -1.55 & - \\

\midrule
 EWC$_\text{ PNAS'17}$~\cite{kirkpatrick2017overcoming} & \xmark & & 50.59 & -42.62&  - \\ 
 LwF$_\text{ TPAMI'17}$~\cite{li2017learning} &  \xmark & & 60.94 & -13.53& 50.05  \\ 
 DyTox$_\text{ CVPR'22}$\cite{douillard2022dytox} & \cmark & \textit{No buffer} &  51.27 & -45.85& 50.46 \\ 
 L2P$_\text{ CVPR'22}$\cite{wang2022learning} & \cmark & & 61.28 & -9.23& 57.34 \\
 S-liPrompts$_\text{ NeurIPS'22}$\cite{wang2022sprompts} & \cmark &  & \textbf{88.65}& \textbf{-0.69}& 76.79 \\
 \rowcolor{mygray} \bf \MoP (ours) & \cmark & & 88.54 & -0.79& \textbf{82.02} \\
\bottomrule
\end{tabular}
}
\end{center}

\vspace{-0.6cm}
\end{table}

\begin{table}[h]
\caption{\textbf{Results on DomainNet for both ID (AA metric) and OOD (CA metric) scenarios}. Best values are highlighted in \textbf{bold}.}
\vspace{-.6cm}
\label{table:domainnetresults}
\begin{center}
\resizebox{0.99\linewidth}{!}{
\begin{tabular}{lcccc}
\toprule 
 Method & Prompt & Buffer size &  \begin{tabular}[c]{@{}c@{}}Seen\\Domains\end{tabular} &  \begin{tabular}[c]{@{}c@{}}Unseen\\Domains\end{tabular} \\
\midrule
 DyTox$_\text{ CVPR'22}$\cite{douillard2022dytox} & \cmark &\textit{50ex/class} & 62.9 \\
\midrule
 DyTox$_\text{ CVPR'22}$\cite{douillard2022dytox} & \cmark & &13.5 & 4.2  \\
 LwF$_\text{ TPAMI'17}$~\cite{li2017learning} & \xmark & & 49.2 & 43.4 \\
CaSSLe$_\text{ CVPR'22}$\cite{fini2022self}(SimCLR \cite{chen2020simple}) &  \xmark& & 48.1 & 45.4 \\
CaSSLe$_\text{ CVPR'22}$\cite{fini2022self}(BYOL \cite{grill2020bootstrap}) & \xmark& & 52.9 & 48.7\\
CaSSLe$_\text{ CVPR'22}$\cite{fini2022self}(Barlow Twins\cite{zbontar2021barlow}) & \xmark&\textit{No buffer} &  51.4 &  47.6 \\
CaSSLe$_\text{ CVPR'22}$\cite{fini2022self}(SupCon~\cite{khosla2020supervised}) & \xmark& & 54.2 & 50.5 \\
 L2P$_\text{ CVPR'22}$\cite{wang2022learning} & \cmark& & 40.1 & 25.5 \\
 S-liPrompts$_\text{ NeurIPS'22}$\cite{wang2022sprompts} & \cmark& & 67.7 & 66.4 \\
 \rowcolor{mygray} \bf \MoP (Ours) & \cmark& & \textbf{69.7}& \textbf{67.0} \\
\bottomrule
\end{tabular}
}
\end{center}
\end{table}

\noindent \textbf{Performance under domain distributional shift.} We now want to assess the benefits of the proposed approach when the testing dataset presents a distributional drift over the training data. 
In particular, we advocated that the proposed approach is a generalization of \cite{wang2022sprompts} to be able to handle samples coming from an unseen distribution. To support this claim, and to demonstrate the superiority of our approach on unseen domains, we resort to the OOD experiments, which are reported in the right-most columns of Tables \ref{table:cddbresults} and \ref{table:domainnetresults}, as well as Table \ref{table:core50results}. From these results, we can observe that excluding S-liPrompts, the performance gains brought by the proposed approach are substantial compared to other \textit{exemplar-free} methods, ranging from 17\% (EWC in CORe50) to 40\% (L2P \cite{wang2022learning} in DomainNet). Even when comparing to state-of-the-art competitors that store exemplars (e.g., DyTox \cite{douillard2022dytox} or Co$^2$L \cite{cha2021co2l} in CORe50), \MoP yields considerable improvements, ranging from 11\% to nearly 17\%. The clear superiority of our approach lies on the isolation of different domains during learning, which do not degenerate the generalization capabilities brought by the pre-trained transformers. Furthermore, when comparing the proposed \MoP to S-liPrompts \cite{wang2022sprompts}, we observe that our method outperforms the latter by around 6\%, 2\% and 3\% in CDDB-Hard, DomainNet and CORe50 benchmarks, respectively. These performance gains on OOD samples might likely come from the flexibility of \MoP in selecting a subset of similar domains for a given test sample, which allows the model to properly weight the contribution of each domain prompt. In contrast, S-liPrompts \cite{wang2022sprompts} forces the model to select only one domain from the seen domains, which impedes its scalability to novel distributions, as empirically shown in these results, as well as in Figure \ref{fig:motivation}.

\begin{table}[t!]
\footnotesize
\centering
\caption{\textbf{Results on CORe50.} Note that CORe50 already provides separate training and testing domains, and thus results can only be computed on the \textbf{OOD scenario}. Results are reported as the Acc metric, where the best values are highlighted in \textbf{bold}. In our method, we use the same $q$ as in the other datasets, whereas * indicates that $q$ is fixed based on the validation set of CORe50, as typically done in all the other approaches. }
\label{table:core50results}
\resizebox{0.99\linewidth}{!}{
\begin{tabular}{lccc}
\toprule 
 Method & Prompt & Buffer size & AA \\
\toprule
GDumb$_\text{ ECCV'20}$~\cite{prabhu2020gdumb} & \xmark && 74.92 \\
BiC$_\text{ CVPR'19}$~\cite{wu2019large} & \xmark& & 79.28 \\
DER++$_\text{ NeurIPS'20}$~\cite{buzzega2020dark} & \xmark& \textit{50ex/class} & 79.70 \\
Co$^2$L$_\text{ ICCV'21}$~\cite{cha2021co2l} &\xmark&& 79.75 \\
DyTox$_\text{ CVPR'22}$ \cite{douillard2022dytox} &\cmark&& 79.21 \\
L2P$_\text{ CVPR'22}$ \cite{wang2022learning} &\cmark& & 81.07 \\
\midrule
EWC$_\text{ PNAS'17}$~\cite{kirkpatrick2017overcoming} & \xmark& & 74.82 \\
LwF$_\text{ TPAMI'17}$~\cite{li2017learning} &\xmark& & 75.45  \\
L2P$_\text{ CVPR'22}$ \cite{wang2022learning} & \cmark& \textit{No buffer} & 78.33 \\

S-liPrompts$_\text{ NeurIPS'22}$ \cite{wang2022sprompts} & \cmark& & 89.06  \\
\rowcolor{mygray} \textbf{\MoP (Ours)} &\cmark& & \textbf{91.43}  \\
\rowcolor{mygray} \textbf{\MoP (Ours)*} & \cmark&& \textbf{92.29}  \\

\bottomrule
\end{tabular}
}
\end{table}

\noindent \textbf{On the impact of the different components.} The empirical study in Table \ref{table:ablationstudy} justifies the need of employing the proposed approach over the strong baseline S-liPrompts \cite{wang2022sprompts}, as well as showcases the impact of each choice.
In a practical scenario, it is unrealistic to assume that the test samples always follow the same distribution as the data used for adaptation. Furthermore, the domain of each sample typically remains unknown. Thus, to align with real-world conditions, we will consider the average of in-distribution and out-of-distribution performance as our metric of reference to evaluate the impact of the different choices. We can observe that in nearly all the cases, the use of an ensembling strategy results in consistent improvements over the single model predictions (considering same distances). An interesting observation is that distances related to the L$_2$-norm typically degrade the performance on ID samples. We observe that in this scenario, the distributions overlap considerably and $p(s|\xx)$ (derived from the Gaussian mixture) is too far from 1 for most ID samples, making the discrimination of samples by these distance measures difficult. Nevertheless, this behavior is reversed in the presence of OOD samples. In particular, our simplification assumes an isotropic Gaussian distribution of the points around the prototypes and therefore reduces the noise in the coordinate-wise variances (which can explain the performance degradation observed when using the Mahanalobis distance), replacing it with distance-wise variances. Thus, the proposed approach combines the best of both worlds, leading to the best average performance across all the configurations.

\begin{table}[h!]
\caption{\textbf{Impact of each design choice of \MoP.} \textit{Maha} denotes the Mahanalobis distance, whereas GMM is used for a Gaussian Mixture Model. Furthermore, \textit{Hybrid} denotes the nature of our approach, which uses an ensembling for OOD samples and a single domain prompt for ID samples. Results (on CDDB-Hard) show the average accuracy (AA), with the deviation from the baseline S-liPrompts \cite{wang2022sprompts} in brackets. Best results in \textbf{bold}.
}
\label{table:ablationstudy}
\centering
\captionsetup{justification=centering}
\resizebox{0.99\linewidth}{!}{
\begin{tabular}{lcccccc}
\toprule 
Method & Ensembling & Distance & \begin{tabular}[c]{@{}c@{}}Seen\\Domains\end{tabular} & \begin{tabular}[c]{@{}c@{}}Unseen\\Domains\end{tabular} & Mean\\ 
\midrule 
S-liPrompts \cite{wang2022sprompts} & \xmark & L1 & 88.65 & 76.79 & 82.72\\
\midrule
\MoP - no ens. (a) & 
\xmark & L2 &  \textbf{89.48} & 76.95 & 83.22$_{(+0.50)}$ \textcolor{blue}{$\uparrow$}\\ 
- & \xmark & Maha & 80.45 & 76.66 & 78.56$_{(-4.16)}$\textcolor{red}{$\downarrow$}\\ 
- & \xmark & L2-GMM &  75.72 & 75.76 & 75.74$_{(-6.98)}$\textcolor{red}{$\downarrow$}\\ 
\midrule
- & \cmark & Uniform & 67.55 & 83.61 &75.58$_{(-7.14)}$ \textcolor{red}{$\downarrow$}\\ 
- & \cmark & L1 &  89.29 & 80.05 & 84.67$_{(+1.95)}$\textcolor{blue}{$\uparrow$}\\ 
- & \cmark & L2 &  68.37 & 84.07 & 76.22$_{(-6.50)}$\textcolor{red}{$\downarrow$}\\ 
- & \cmark & Maha & 80.48 & 77.56 & 79.02$_{(-3.70)}$\textcolor{red}{$\downarrow$}\\ 
\MoP - ens. (b) & \cmark & L2-GMM   & 72.51 & \textbf{89.21} & 80.86$_{(-1.86)}$\textcolor{red}{$\downarrow$} \\ 
\midrule
\textbf{\MoP (Proposed)} & Hybrid & ID (a)/ OOD (b) & 88.54 & 82.02 & \textbf{85.28}$_{(+2.56)}$\textcolor{blue}{$\uparrow$}\\ 
\bottomrule
\end{tabular}
}
\end{table}

\begin{figure}[h!]
    \centering
    \includegraphics[width=0.95\linewidth]{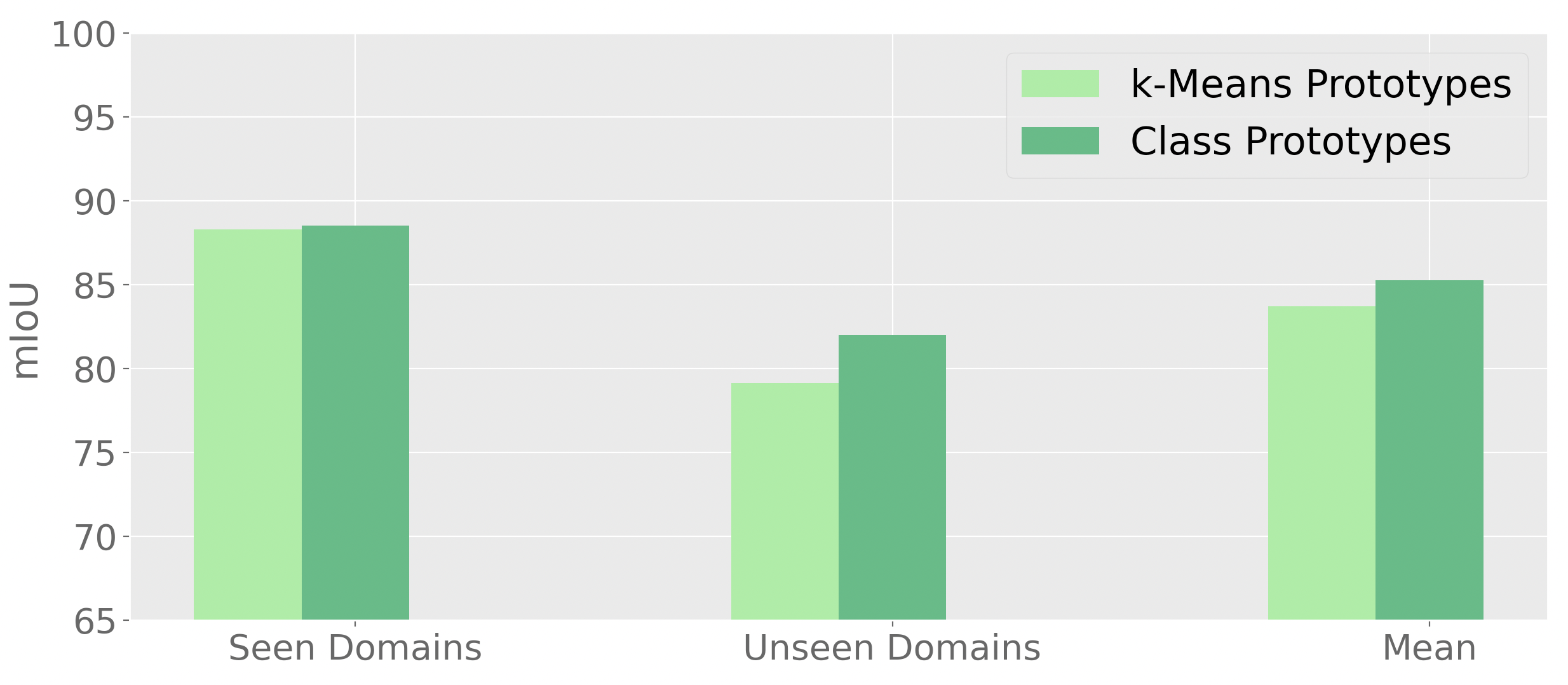}
    \caption{\textbf{k-Means or class prototypes as domain centroids?} Ablation study that demonstrates the benefits of using class prototypes (our approach) rather than k-Means prototypes, as in \cite{wang2022sprompts}.}
    \label{fig:centroids}
\end{figure}

\label{strategy_prompts}
\noindent \textbf{Strategy to select the domain prompts.} 
As emphasized in Sec.~\ref{sec:infer}, \cite{wang2022sprompts} uses K-Means over the features extracted with a pre-trained ViT to compute the prototypes which are used to dynamically select which prompt to use at test time. While this strategy is memory efficient, it lacks flexibility, as the number of clusters needs to be adjusted according to the dataset employed. To alleviate this issue, we instead use class-wise prototypes as a \textit{hyperparameter-free} alternative to compute representative prototypes. The effect of using either k-Means or class-prototypes is depicted in Fig.~\ref{fig:centroids}. From these results, we empirically observe that this choice
improves performance in both in-distribution and out-of-distribution domains, leading to a higher average performance. Furthermore, it is noteworthy to mention that using class-wise prototypes makes the distribution of points around prototypes Gaussian, which explains the satisfactory performance of MoP-CLIP, particularly on samples from unseen domains.

\begin{figure}[h!]
    \centering
    \includegraphics[width=1.0\linewidth]{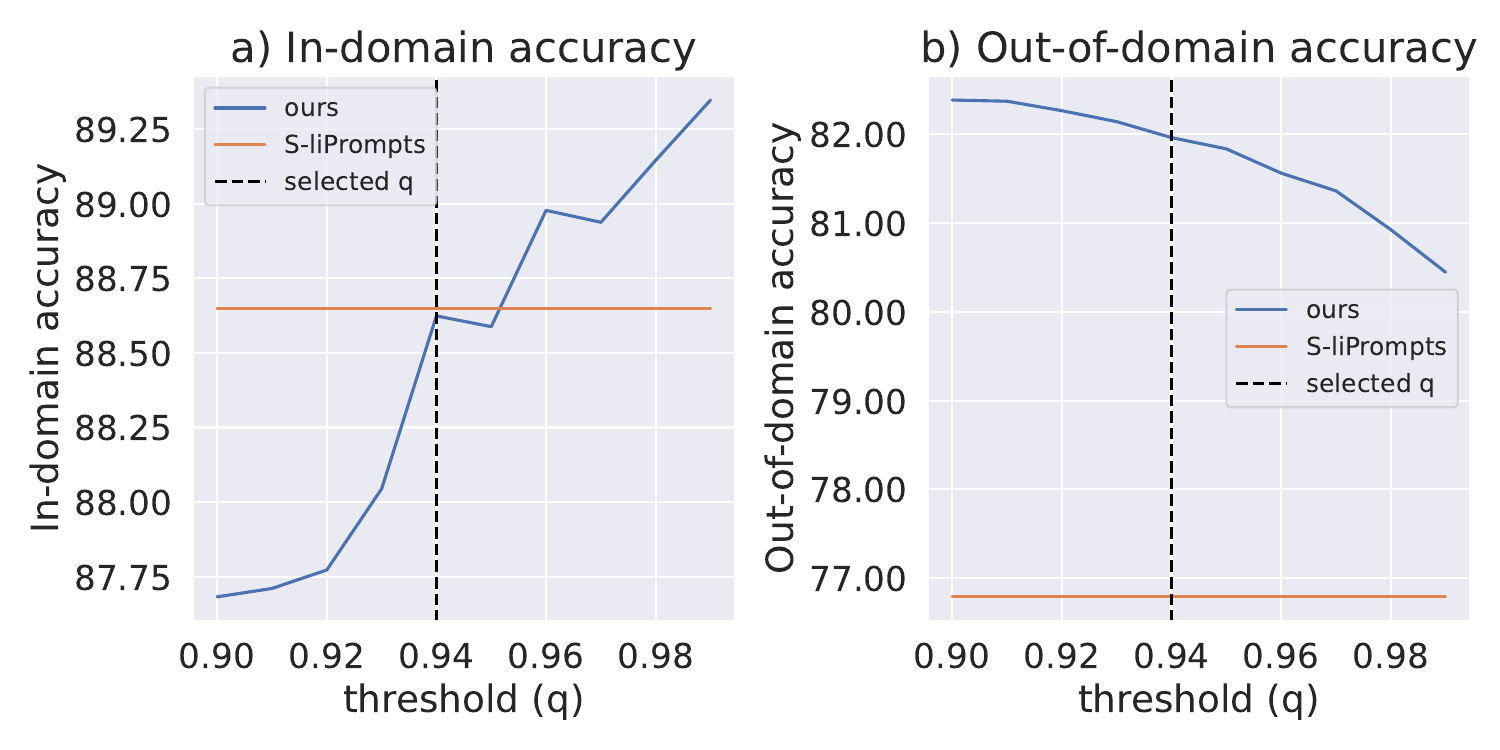}
    \caption{\textbf{A controllable trade-off between in-domain and out-of-domain prediction performances.} Impact of the threshold $q$ (Sec. \ref{sec:infer}) on the accuracy, evaluated on  CDDB-Hard.}
    \label{fig:ablation-study}
\end{figure}

\noindent \textbf{How much trade-off is sufficient?} The influence of the threshold $q$ from our simple out-of-distribution criterion (Sec. \ref{domain_select}) to select between seen and unseen domains is shown in Figure \ref{fig:ablation-study}. As stressed earlier, we aim for a compromise between ID and OOD performance, in order to provide generalizable models. As target domains should remain unknown at inference, we selected a fixed $q$ value that provided the optimal average performance across both settings. Nevertheless, these plots reveal two interesting findings. First, the average performance of the model is not very sensitive to the choice of $q$. For example, the performance of ID samples decreases as $q$ decreases, whereas OOD performance improves. On the other hand, if $q$ increases, the accuracy in the ID scenario increases, while it decreases for OOD samples. And second, if prior knowledge about the target domain is available --an assumption made by all existing DIL literature-- the performance of \MoP is further increased, enlarging the gap with SOTA methods.

\section{Conclusion}
Findings from this work reveal that existing literature on domain incremental learning suffers under the presence of distributional drift, hampering their scalability to practical scenarios. To overcome this issue, we have proposed a generalization of the recent S-ilPrompts \cite{wang2022sprompts} approach, that further handles out-of-distribution samples. In addition to outperforming current state-of-the-art, particularly in the unseen domain setting, our method brings several interesting benefits compared to most existing DIL method. 
First, \MoP is \textit{exemplar-free}, eliminating the limitations of conventional DIL approaches in terms of storage and privacy. Furthermore, as prompts are learned independently on each domain, and the model parameters remain fixed during the adaptation, the performance of our approach is insensitive to the ordering of the seen domains.  This contrasts with a whole body of the literature, where the choice of the sequence order can significantly impact the final performance. Our comprehensive evaluation shows the empirical gains provided by MoP-CLIP, pointing to visual prompt tuning as an appealing alternative for general domain incremental learning. Finally, we stress that while powerful, the proposed approach retains the spirit of S-ilPrompts \cite{wang2022sprompts}, which advocates for a simple yet elegant method.

\noindent \textit{Potential Negative Impact:} Language-vision models and prompt tuning heavily rely on pre-training data, including different corpus, which may contain biases and reinforce existing societal prejudices. The use of text prompt tuning might amplify these biases and contribute to biased classification results.

\def\faChecked{\FA\symbol{"F00C}}
\def\faCrossed{\FA\symbol{"F00D}}


{\small
\bibliographystyle{ieee_fullname}
\bibliography{egbib}

\begin{thebibliography}{10}\itemsep=-1pt

\bibitem{ahn2021ss}
Hongjoon Ahn, Jihwan Kwak, Subin Lim, Hyeonsu Bang, Hyojun Kim, and Taesup
  Moon.
\newblock Ss-il: Separated softmax for incremental learning.
\newblock In {\em Proceedings of the IEEE/CVF International conference on
  computer vision}, pages 844--853, 2021.

\bibitem{aljundi2019gradient}
Rahaf Aljundi, Min Lin, Baptiste Goujaud, and Yoshua Bengio.
\newblock Gradient based sample selection for online continual learning.
\newblock {\em Advances in neural information processing systems}, 32, 2019.

\bibitem{bang2021rainbow}
Jihwan Bang, Heesu Kim, YoungJoon Yoo, Jung-Woo Ha, and Jonghyun Choi.
\newblock Rainbow memory: Continual learning with a memory of diverse samples.
\newblock In {\em Proceedings of the IEEE/CVF Conference on Computer Vision and
  Pattern Recognition}, pages 8218--8227, 2021.

\bibitem{brown2020language}
Tom Brown, Benjamin Mann, Nick Ryder, Melanie Subbiah, Jared~D Kaplan, Prafulla
  Dhariwal, Arvind Neelakantan, Pranav Shyam, Girish Sastry, Amanda Askell,
  et~al.
\newblock Language models are few-shot learners.
\newblock {\em Advances in neural information processing systems},
  33:1877--1901, 2020.

\bibitem{buzzega2020dark}
Pietro Buzzega, Matteo Boschini, Angelo Porrello, Davide Abati, and Simone
  Calderara.
\newblock Dark experience for general continual learning: a strong, simple
  baseline.
\newblock In {\em NeurIPS}, 2020.

\bibitem{cha2021co2l}
Hyuntak Cha, Jaeho Lee, and Jinwoo Shin.
\newblock Co2l: Contrastive continual learning.
\newblock In {\em Proceedings of the IEEE/CVF International conference on
  computer vision}, pages 9516--9525, 2021.

\bibitem{chaudhry2018riemannian}
Arslan Chaudhry, Puneet~K Dokania, Thalaiyasingam Ajanthan, and Philip~HS Torr.
\newblock Riemannian walk for incremental learning: Understanding forgetting
  and intransigence.
\newblock In {\em Proceedings of the European conference on computer vision
  (ECCV)}, pages 532--547, 2018.

\bibitem{chaudhry2019tiny}
Arslan Chaudhry, Marcus Rohrbach, Mohamed Elhoseiny, Thalaiyasingam Ajanthan,
  Puneet~K Dokania, Philip~HS Torr, and Marc'Aurelio Ranzato.
\newblock On tiny episodic memories in continual learning.
\newblock {\em arXiv preprint arXiv:1902.10486}, 2019.

\bibitem{chen2023understanding}
Aochuan Chen, Yuguang Yao, Pin-Yu Chen, Yihua Zhang, and Sijia Liu.
\newblock Understanding and improving visual prompting: A label-mapping
  perspective.
\newblock In {\em Proceedings of the IEEE/CVF Conference on Computer Vision and
  Pattern Recognition}, pages 19133--19143, 2023.

\bibitem{chen2022compound}
Chaoqi Chen, Jiongcheng Li, Xiaoguang Han, Xiaoqing Liu, and Yizhou Yu.
\newblock Compound domain generalization via meta-knowledge encoding.
\newblock In {\em Proceedings of the IEEE/CVF Conference on Computer Vision and
  Pattern Recognition}, pages 7119--7129, 2022.

\bibitem{chen2020simple}
Ting Chen, Simon Kornblith, Mohammad Norouzi, and Geoffrey Hinton.
\newblock A simple framework for contrastive learning of visual
  representations.
\newblock In {\em International conference on machine learning}, pages
  1597--1607. PMLR, 2020.

\bibitem{dosovitskiy2020image}
Alexey Dosovitskiy, Lucas Beyer, Alexander Kolesnikov, Dirk Weissenborn,
  Xiaohua Zhai, Thomas Unterthiner, Mostafa Dehghani, Matthias Minderer, Georg
  Heigold, Sylvain Gelly, et~al.
\newblock An image is worth 16x16 words: Transformers for image recognition at
  scale.
\newblock In {\em International Conference on Learning Representations}, 2021.

\bibitem{douillard2022dytox}
Arthur Douillard, Alexandre Ram{\'e}, Guillaume Couairon, and Matthieu Cord.
\newblock Dytox: Transformers for continual learning with dynamic token
  expansion.
\newblock In {\em Proceedings of the IEEE/CVF Conference on Computer Vision and
  Pattern Recognition}, pages 9285--9295, 2022.

\bibitem{fini2022self}
Enrico Fini, Victor G~Turrisi Da~Costa, Xavier Alameda-Pineda, Elisa Ricci,
  Karteek Alahari, and Julien Mairal.
\newblock Self-supervised models are continual learners.
\newblock In {\em Proceedings of the IEEE/CVF Conference on Computer Vision and
  Pattern Recognition}, pages 9621--9630, 2022.

\bibitem{grill2020bootstrap}
Jean-Bastien Grill, Florian Strub, Florent Altch{\'e}, Corentin Tallec, Pierre
  Richemond, Elena Buchatskaya, Carl Doersch, Bernardo Avila~Pires, Zhaohan
  Guo, Mohammad Gheshlaghi~Azar, et~al.
\newblock Bootstrap your own latent-a new approach to self-supervised learning.
\newblock {\em Advances in neural information processing systems},
  33:21271--21284, 2020.

\bibitem{hou2018lifelong}
Saihui Hou, Xinyu Pan, Chen~Change Loy, Zilei Wang, and Dahua Lin.
\newblock Lifelong learning via progressive distillation and retrospection.
\newblock In {\em Proceedings of the European Conference on Computer Vision
  (ECCV)}, pages 437--452, 2018.

\bibitem{hou2019learning}
Saihui Hou, Xinyu Pan, Chen~Change Loy, Zilei Wang, and Dahua Lin.
\newblock Learning a unified classifier incrementally via rebalancing.
\newblock In {\em Proceedings of the IEEE/CVF conference on Computer Vision and
  Pattern Recognition}, pages 831--839, 2019.

\bibitem{jia2022visual}
Menglin Jia, Luming Tang, Bor-Chun Chen, Claire Cardie, Serge Belongie, Bharath
  Hariharan, and Ser-Nam Lim.
\newblock Visual prompt tuning.
\newblock In {\em Computer Vision--ECCV 2022: 17th European Conference, Tel
  Aviv, Israel, October 23--27, 2022, Proceedings, Part XXXIII}, pages
  709--727. Springer, 2022.

\bibitem{ju2022prompting}
Chen Ju, Tengda Han, Kunhao Zheng, Ya Zhang, and Weidi Xie.
\newblock Prompting visual-language models for efficient video understanding.
\newblock In {\em Computer Vision--ECCV 2022: 17th European Conference, Tel
  Aviv, Israel, October 23--27, 2022, Proceedings, Part XXXV}, pages 105--124.
  Springer, 2022.

\bibitem{khosla2020supervised}
Prannay Khosla, Piotr Teterwak, Chen Wang, Aaron Sarna, Yonglong Tian, Phillip
  Isola, Aaron Maschinot, Ce Liu, and Dilip Krishnan.
\newblock Supervised contrastive learning.
\newblock {\em Advances in neural information processing systems},
  33:18661--18673, 2020.

\bibitem{kirkpatrick2017overcoming}
James Kirkpatrick, Razvan Pascanu, Neil Rabinowitz, Joel Veness, Guillaume
  Desjardins, Andrei~A Rusu, Kieran Milan, John Quan, Tiago Ramalho, Agnieszka
  Grabska-Barwinska, et~al.
\newblock Overcoming catastrophic forgetting in neural networks.
\newblock {\em Proceedings of the national academy of sciences},
  114(13):3521--3526, 2017.

\bibitem{le2021many}
Teven Le~Scao and Alexander~M Rush.
\newblock How many data points is a prompt worth?
\newblock In {\em Proceedings of the 2021 Conference of the North American
  Chapter of the Association for Computational Linguistics: Human Language
  Technologies}, pages 2627--2636, 2021.

\bibitem{lee2019overcoming}
Kibok Lee, Kimin Lee, Jinwoo Shin, and Honglak Lee.
\newblock Overcoming catastrophic forgetting with unlabeled data in the wild.
\newblock In {\em Proceedings of the IEEE/CVF International Conference on
  Computer Vision}, pages 312--321, 2019.

\bibitem{li2023continual}
Chuqiao Li, Zhiwu Huang, Danda~Pani Paudel, Yabin Wang, Mohamad Shahbazi,
  Xiaopeng Hong, and Luc Van~Gool.
\newblock A continual deepfake detection benchmark: Dataset, methods, and
  essentials.
\newblock In {\em Proceedings of the IEEE/CVF Winter Conference on Applications
  of Computer Vision}, pages 1339--1349, 2023.

\bibitem{li2017learning}
Zhizhong Li and Derek Hoiem.
\newblock Learning without forgetting.
\newblock {\em IEEE transactions on pattern analysis and machine intelligence},
  40(12):2935--2947, 2017.

\bibitem{liu2022deja}
Chenxi Liu, Lixu Wang, Lingjuan Lyu, Chen Sun, Xiao Wang, and Qi Zhu.
\newblock Deja vu: Continual model generalization for unseen domains.
\newblock In {\em International Conference on Learning Representations}, 2023.

\bibitem{core50}
Vincenzo Lomonaco and Davide Maltoni.
\newblock Core50: a new dataset and benchmark for continuous object
  recognition.
\newblock {\em CoRR}, abs/1705.03550, 2017.

\bibitem{lopez2017gradient}
David Lopez-Paz and Marc'Aurelio Ranzato.
\newblock Gradient episodic memory for continual learning.
\newblock {\em Advances in neural information processing systems}, 30, 2017.

\bibitem{lu2022prompt}
Yuning Lu, Jianzhuang Liu, Yonggang Zhang, Yajing Liu, and Xinmei Tian.
\newblock Prompt distribution learning.
\newblock In {\em Proceedings of the IEEE/CVF Conference on Computer Vision and
  Pattern Recognition}, pages 5206--5215, 2022.

\bibitem{marra2019incremental}
Francesco Marra, Cristiano Saltori, Giulia Boato, and Luisa Verdoliva.
\newblock Incremental learning for the detection and classification of
  gan-generated images.
\newblock In {\em 2019 IEEE international workshop on information forensics and
  security (WIFS)}, pages 1--6. IEEE, 2019.

\bibitem{pellegrini2020latent}
Lorenzo Pellegrini, Gabriele Graffieti, Vincenzo Lomonaco, and Davide Maltoni.
\newblock Latent replay for real-time continual learning.
\newblock In {\em 2020 IEEE/RSJ International Conference on Intelligent Robots
  and Systems (IROS)}, pages 10203--10209. IEEE, 2020.

\bibitem{domainnet}
Xingchao Peng, Qinxun Bai, Xide Xia, Zijun Huang, Kate Saenko, and Bo Wang.
\newblock Moment matching for multi-source domain adaptation.
\newblock In {\em Proceedings of the IEEE/CVF international conference on
  computer vision}, pages 1406--1415, 2019.

\bibitem{prabhu2020gdumb}
Ameya Prabhu, Philip~HS Torr, and Puneet~K Dokania.
\newblock Gdumb: A simple approach that questions our progress in continual
  learning.
\newblock In {\em ECCV}, 2020.

\bibitem{rolnick2019experience}
David Rolnick, Arun Ahuja, Jonathan Schwarz, Timothy Lillicrap, and Gregory
  Wayne.
\newblock Experience replay for continual learning.
\newblock {\em Advances in Neural Information Processing Systems}, 32, 2019.

\bibitem{imagenet}
Olga Russakovsky, Jia Deng, Hao Su, Jonathan Krause, Sanjeev Satheesh, Sean Ma,
  Zhiheng Huang, Andrej Karpathy, Aditya Khosla, Michael Bernstein, et~al.
\newblock Imagenet large scale visual recognition challenge.
\newblock {\em International journal of computer vision}, 115:211--252, 2015.

\bibitem{shin2017continual}
Hanul Shin, Jung~Kwon Lee, Jaehong Kim, and Jiwon Kim.
\newblock Continual learning with deep generative replay.
\newblock {\em Advances in neural information processing systems}, 30, 2017.

\bibitem{sohn2023visual}
Kihyuk Sohn, Huiwen Chang, Jos{\'e} Lezama, Luisa Polania, Han Zhang, Yuan Hao,
  Irfan Essa, and Lu Jiang.
\newblock Visual prompt tuning for generative transfer learning.
\newblock In {\em Proceedings of the IEEE/CVF Conference on Computer Vision and
  Pattern Recognition}, pages 19840--19851, 2023.

\bibitem{wang2022sprompts}
Yabin Wang, Zhiwu Huang, and Xiaopeng Hong.
\newblock S-prompts learning with pre-trained transformers: An
  occam{\textquoteright}s razor for domain incremental learning.
\newblock In Alice~H. Oh, Alekh Agarwal, Danielle Belgrave, and Kyunghyun Cho,
  editors, {\em Advances in Neural Information Processing Systems}, 2022.

\bibitem{wang2021learning}
Zijian Wang, Yadan Luo, Ruihong Qiu, Zi Huang, and Mahsa Baktashmotlagh.
\newblock Learning to diversify for single domain generalization.
\newblock In {\em Proceedings of the IEEE/CVF International Conference on
  Computer Vision}, pages 834--843, 2021.

\bibitem{wang2022learning}
Zifeng Wang, Zizhao Zhang, Chen-Yu Lee, Han Zhang, Ruoxi Sun, Xiaoqi Ren,
  Guolong Su, Vincent Perot, Jennifer Dy, and Tomas Pfister.
\newblock Learning to prompt for continual learning.
\newblock In {\em Proceedings of the IEEE/CVF Conference on Computer Vision and
  Pattern Recognition}, pages 139--149, 2022.

\bibitem{wu2019large}
Yue Wu, Yinpeng Chen, Lijuan Wang, Yuancheng Ye, Zicheng Liu, Yandong Guo, and
  Yun Fu.
\newblock Large scale incremental learning.
\newblock In {\em CVPR}, 2019.

\bibitem{xing2022class}
Yinghui Xing, Qirui Wu, De Cheng, Shizhou Zhang, Guoqiang Liang, and Yanning
  Zhang.
\newblock Class-aware visual prompt tuning for vision-language pre-trained
  model.
\newblock {\em arXiv preprint arXiv:2208.08340}, 2022.

\bibitem{yao2022pcl}
Xufeng Yao, Yang Bai, Xinyun Zhang, Yuechen Zhang, Qi Sun, Ran Chen, Ruiyu Li,
  and Bei Yu.
\newblock Pcl: Proxy-based contrastive learning for domain generalization.
\newblock In {\em Proceedings of the IEEE/CVF Conference on Computer Vision and
  Pattern Recognition}, pages 7097--7107, 2022.

\bibitem{zbontar2021barlow}
Jure Zbontar, Li Jing, Ishan Misra, Yann LeCun, and St{\'e}phane Deny.
\newblock Barlow twins: Self-supervised learning via redundancy reduction.
\newblock In {\em International Conference on Machine Learning}, pages
  12310--12320. PMLR, 2021.

\bibitem{zenke2017continual}
Friedemann Zenke, Ben Poole, and Surya Ganguli.
\newblock Continual learning through synaptic intelligence.
\newblock In {\em International conference on machine learning}, pages
  3987--3995. PMLR, 2017.

\bibitem{zhang2023nico++}
Xingxuan Zhang, Yue He, Renzhe Xu, Han Yu, Zheyan Shen, and Peng Cui.
\newblock Nico++: Towards better benchmarking for domain generalization.
\newblock In {\em Proceedings of the IEEE/CVF Conference on Computer Vision and
  Pattern Recognition}, pages 16036--16047, 2023.

\bibitem{zhang2022exact}
Yabin Zhang, Minghan Li, Ruihuang Li, Kui Jia, and Lei Zhang.
\newblock Exact feature distribution matching for arbitrary style transfer and
  domain generalization.
\newblock In {\em Proceedings of the IEEE/CVF Conference on Computer Vision and
  Pattern Recognition}, pages 8035--8045, 2022.

\bibitem{zhou2022conditional}
Kaiyang Zhou, Jingkang Yang, Chen~Change Loy, and Ziwei Liu.
\newblock Conditional prompt learning for vision-language models.
\newblock In {\em Proceedings of the IEEE/CVF Conference on Computer Vision and
  Pattern Recognition}, pages 16816--16825, 2022.

\bibitem{zhou2022learning}
Kaiyang Zhou, Jingkang Yang, Chen~Change Loy, and Ziwei Liu.
\newblock Learning to prompt for vision-language models.
\newblock {\em International Journal of Computer Vision}, 130(9):2337--2348,
  2022.

\end{thebibliography}
}

\setcounter{equation}{0}

\section*{Supplementary Material}

\section{Proof of the proposed approximation}
The details of how we obtain the model presented in equation (5) in the main paper can be found below:
\setcounter{equation}{0}
\begin{equation}
\begin{split}
 w_s & = p(\xx \in \mathcal{D}_{s})\\
    & = p(s' = s | \xx)\\
    & = \frac{p(\xx|s) \cdot p(s)}{p(\xx)} \text{ (Bayes theorem)}  \\
    & = \frac{p(\xx|s) \cdot p(s)}{\sum_j p(\xx|j) \cdot p(j)} \text{ (Marginalization)} \\
    & = \frac{p(\xx|s)}{\sum_j p(\xx|j)} \text{ ($\mathcal{H}_1)$} \\    
    & = \frac{p(\Delta_s(\xx)|s)}{\sum_j p(\Delta_j(\xx)|j)} \text{ (}\mathcal{H}_2 \text{)} \\
     & = \frac{\mathcal{N}(\Delta_s(\xx);\mu^{k^*}_{s}, \sigma^{k^*}_{s})}{\sum_j \mathcal{N}(\Delta_j(\xx);\mu^{t^*}_{j}, \sigma^{t^*}_{j})}, \\
\end{split}
\label{ensembling-supp}
\end{equation}

We have to make three assumptions or hypothesis to derive this model:
\begin{itemize}
\item $\mathcal{H}_1$: Each domain is of equal importance in our scenario, i.e. if we consider the probability of the sample belonging to a certain domain uniform when we have no a priori on the sample. 
\item $\mathcal{H}_2$: $p(\xx|s) \approx p(\Delta_s(\xx)|s)$, i.e. the distribution of $f_{\theta}(\xx_{\mr{tok}})$ with $x_{\mr{tok}} \in \mathcal{D}_s$ is isotropic. 
\item $\mathcal{H}_3$: $\Delta_s(\boldsymbol{x})|s \sim \mathcal{N}(\cdot ; \mu^{k^*}_{s}, \sigma^{k^*}_{s})$, i.e. $\boldsymbol{x})|s$ follows a Gaussian of mean $\mu^{k^*}_s$ and standard deviation $\sigma^{k^*}_s$.
\end{itemize}

$\mathcal{H}_1$ is reasonable in practice as test sample can come from any domain with equal probability. $\mathcal{H}_2$ and $\mathcal{H}_3$ are made to simplify the model, make it easy to store in memory and to compute. These hypothesis transform the mixture weights model into a Gaussian Mixture Model on the distances to the prototypes (L2-GMM).
Please note that in our case the ensembling with the Mahanalobis distance is equivalent to the well known classical GMM using directly the features and the prototypes to derive $p(\xx \in \mathcal{D}_{s})$.

We empirically observe in the ablation study (Table (4) in the main paper) that the usage of this Gaussian Mixture Model on the distances to the prototypes yields superior performance compared to a GMM using directly the features and the prototypes. We suspect that these approximations are efficient because they reduce the coordinate-wise noise in the standard deviations inherent to the Mahanalobis distance. Gaussian seems like a good approximation of $\Delta_s(\boldsymbol{x})|s$, even though the approximation using other distributions could be investigated in the future, such as the Weibull Distribution or the Generalized Pareto Distribution. 

\section{Algorithm}

The detailed algorithm of the proposed \MoP approach is shown in Algorithm \ref{alg_test}. $\xx$ denotes the samples to be classified, $f_{\theta}$ and $f_{\phi}$ the visual and text encoder of the network and $\mathcal{P}^V$, $\mathcal{P}^{T}$ the sets of visual of text prompts and $\mathcal{E}$ the domains prototypes learned during training.
$\mathcal{G} =\{(\mu_s^k ; \sigma_s^k), s=1..N, k=1..K\}$ denotes the parameters of the Gaussian distributions learned for the different domains $s$ and classes $k$.
\begin{algorithm}
\caption{Inference procedure for the proposed method}
\label{alg_test}
\renewcommand{\algorithmicrequire}{\textbf{Input:}}
\renewcommand{\algorithmicensure}{\textbf{Output:}}
\begin{algorithmic}[1]
\STATE Input: 
$\xx;$
$f_{\theta}; $
$f_{\phi}; $
$\mathcal{P}^V;$
$\mathcal{P}^{T};$
$\mathcal{E};$
$\mathcal{G};$

\STATE Init $E \in O^{K \times N}$
\STATE Compute image features: $f_x \gets f_{\theta}(\xx_{tok})$
\STATE Compute matrix $D$ : $D_{i,j} \gets ||f_x - \boldsymbol{m}_j^i||_2$
\STATE Compute matrix $D'$ : $D'_{j} \gets  \operatorname*{min}_{i} D_{i,j}$
\IF{$F(\Delta_{s^*}(\xx)) \leq q$ ($\xx$ is In-Domain) }
  \STATE $W_{s^*} =1, \forall s \neq s^*, W_{s} =0$.
    \STATE Compute prediction using the best prompt:
    \FOR{$k=1,2,...,K $ } 
    \STATE  $\xx_{pro} \gets [\xx_{tok}, \pp^v_{s^*} ,x_{cls}]$
    \STATE $t_j \gets \left [ \pp^t_{s^*}, c_j \right ]$
    
    \STATE $E_{k,{s^*}} \gets \frac{\exp (cos( f_{\theta}(\xx_{pro}),f_{\phi}(t_k) )  )}{ {\textstyle \sum_{i=1}^{C}}\exp (cos( f_{\theta}(\xx_{pro}),f_{\phi}(t_i) )) }$ 
    \ENDFOR
  \ELSE
  \STATE Compute $W$ using equation (5), $D'$ and $\{(\mu^{k^*}_{s}, \sigma^{k^*}_{s}) \}_{s=1}^N$.
    \STATE Compute predictions using the different prompts:
    \FOR{$s=1,2,...,N $ } 
    \FOR{$k=1,2,...,K $ } 
    \STATE  $\xx_{pro} \gets [\xx_{tok}, \pp^v_s ,x_{cls}]$
    \STATE $t_j \gets \left [ \pp^t_s, c_j \right ]$
    
    \STATE $E_{k,s} \gets \frac{\exp (cos( f_{\theta}(\xx_{pro}),f_{\phi}(t_k) )  )}{ {\textstyle \sum_{i=1}^{C}}\exp (cos( f_{\theta}(\xx_{pro}),f_{\phi}(t_i) )) }$ 
    \ENDFOR
    \ENDFOR

\ENDIF
\STATE $P \gets E \cdot W^{T}$ 
Return P the soft classification vector
\end{algorithmic}
\end{algorithm}
\setcounter{table}{4}

\section{Additional results}
Table \ref{confusiontable} emphasizes that S-Prompts performances degrade when evaluation is done on unseen
domains, and shows that the proposed \MoP seems to generalize better, mitigating the performance degradation under domain distributions. In particular, the left-side section reports the results of S-Prompts trained separately on the different domains (\textit{x-axis}) and evaluated in each of the domains (\textit{y-axis}). For example, $67.41$ denotes the accuracy of the model trained solely on Infograph domain and tested on the Clipart domain. We use \textcolor{blue}{blue} to denote the performance of in-distribution samples (when train and test data are drawn from the same distribution), which can be considered as an upper bound, as there is no distributional drift between samples. Then, both results in black and \textcolor{magenta}{magenta} highlight the results for each tested domain, assuming that the tested domain remains unknown and all training samples come from the same domain (specified in each column). Note that across each test domain we highlight the results from the best model in \textcolor{magenta}{magenta}. If we look at the results obtained by S-Prompts under ID and OOD conditions (\textit{S-Prompts (ID)} and \textit{S-Prompts (OOD)} columns), we can observe that: \textit{i)} its performance deteriorates under domain shift and \textit{ii)}, the selection criterion of S-Prompts is not always optimal.
On the other hand, the proposed approach (\textit{last column}) substantially outperforms S-Prompts in five out of six domains, as well as the best out-of-distribution model (in magenta).

\begin{table*}[t!]
    \centering
    \caption{\textbf{Empirical motivation of resorting to the prediction ensembling scheme for OOD situations.} 
    Classification accuracy across DomainNet domains using different specialized prompts, for both single and ensembling predictions. The results \textcolor{blue}{blue} denote the accuracy with the 
    in-domain prompts, whereas results in \textcolor{magenta}{magenta} denote the accuracy using the best out-of-domain prompts (prompts from all domains except the current one). Furthermore,  results in bold (\textit{last column}) denote the highest accuracy amongst out-of-domain methods. For 5 out of 6 domain sets, the proposed prediction ensembling method yields higher accuracy than the best out-of-domain prompt. This suggests that the ensembling technique is overall relevant when test examples are from a novel domain (i.e. unseen during the training).}

    \label{confusiontable}
    \small{
    \begin{tabular}{l|cccccc|c|c|c}
        \toprule
         & Clipart & Infograph & Painting & Quickdraw & Real  & Sketch & S-Prompts (ID) & S-Prompts (OOD) & Pred. Ens. (OOD) \\
        \midrule
        Clipart & \textcolor{blue}{80.14} & 67.41 & 64.77 & 38.9 & \textcolor{magenta}{69.49} & 69.02 & 78,57 & 69,31 &\textbf{73.48}$_{(+4.01)}$ \\
        Infograph & 44.59 & \textcolor{blue}{60.65} & 43.24 & 15.36 & \textcolor{magenta}{48.93} & 36.08  &  58,72 & 46.50 &\textbf{50.40}$_{(+1.47)}$ \\
        Painting & 59.56 & 61.88 & \textcolor{blue}{78.00} & 24.97 & \textcolor{magenta}{64.43} & 57.32  & 74,76  & 61,88 &\textbf{67.93}$_{(+3.50)}$ \\
        Quickdraw & 16.8 & 13.11 & 8.30 & \textcolor{blue}{46.65} & 13.58 & \textbf{\textcolor{magenta}{17.29}} & 46,59  & 16,79& 16.78$_{(-0.51)}$ \\
        Real & 78.35 & \textcolor{magenta}{79.38} & 75.83 & 45.44 & \textcolor{blue}{87.94} & 71.79 & 85,19 &  77,38 &\textbf{83.48}$_{(+4.10)}$ \\
        Sketch & 61.51 & 59.18 & 55.22 & 30.43 & \textcolor{magenta}{61.59} & \textcolor{blue}{72.97} & 69,76 &  58,87 & \textbf{66.31}$_{(+4.72)}$ \\
        \bottomrule
    \end{tabular}}
\end{table*}

\end{document}